\documentclass[10pt,twocolumn,letterpaper]{article}

\usepackage{wacv}
\usepackage{times}
\usepackage{epsfig}
\usepackage{graphicx}
\usepackage{amsmath}
\usepackage{amssymb}
\usepackage{booktabs}

\usepackage{subcaption}
\usepackage{multirow}
\usepackage[table,xcdraw]{xcolor}
\usepackage{enumitem}

\newcolumntype{L}[1]{>{\raggedright\let\newline\\\arraybackslash\hspace{0pt}}m{#1}}
\newcolumntype{C}[1]{>{\centering\let\newline\\\arraybackslash\hspace{0pt}}m{#1}}
\newcolumntype{R}[1]{>{\raggedleft\let\newline\\\arraybackslash\hspace{0pt}}m{#1}}

\def\wacvPaperID{-} 

\wacvalgorithmstrack   

\wacvfinalcopy 

\ifwacvfinal
\usepackage[breaklinks=true,bookmarks=false]{hyperref}
\else
\usepackage[pagebackref=true,breaklinks=true,colorlinks,bookmarks=false]{hyperref}
\fi

\begin{document}

\title{Learning More May Not Be Better: \\ Knowledge Transferability in Vision and Language Tasks}

\author{Tianwei Chen$^1$, Noa Garcia$^1$, Mayu Otani$^2$, Chenhui Chu$^3$,  Yuta Nakashima$^1$, Hajime Nagahara$^1$\\
Osaka University$^1$, CyberAgent Inc.$^2$, Kyoto University$^3$\\
{\tt\small \{chentw@is., noagarcia@, n-yuta@, nagahara@\}ids.osaka-u.ac.jp}\\
{\tt\small otani\_mayu@cyberagent.co.jp}\\
{\tt\small chu@i.kyoto-u.ac.jp}
}

\maketitle
\thispagestyle{empty}

\begin{abstract}
Is more data always better to train vision-and-language models? We study knowledge transferability in multi-modal tasks. The current tendency in machine learning is to assume that by joining multiple datasets from different tasks their overall performance will improve. However, we show that not all the knowledge transfers well or has a positive impact on related tasks, even when they share a common goal. We conduct an exhaustive analysis based on hundreds of cross-experiments on 12 vision-and-language tasks categorized in 4 groups. Whereas tasks in the same group are prone to improve each other, results show that this is not always the case. Other factors such as dataset size or pre-training stage have also a great impact on how well the knowledge is transferred.
\end{abstract}


\section{Introduction}
\label{sec:intro}

\textit{The more data, the better} seems to be the current \textit{motto} in machine learning, as large language models get exceptional results on previously unseen tasks by being trained on hundreds of millions of samples crawled from the Internet \cite{brown2020language,reed2022generalist,alayrac2022flamingo,ramesh2022hierarchical}. Following the path led by natural language processing research, the computer vision community is gradually adopting Transformer-based models trained on web-scale datasets to achieve high performance in zero-shot settings \cite{FhdcDosovitskiy2021AnII,radford2021learning}. This is conducted by leveraging huge amounts of image-caption pairs available online to let the models learn correspondences between the language semantics and the visual appearance of objects.

\begin{figure}
  \centering
  \includegraphics[width = 0.47\textwidth]{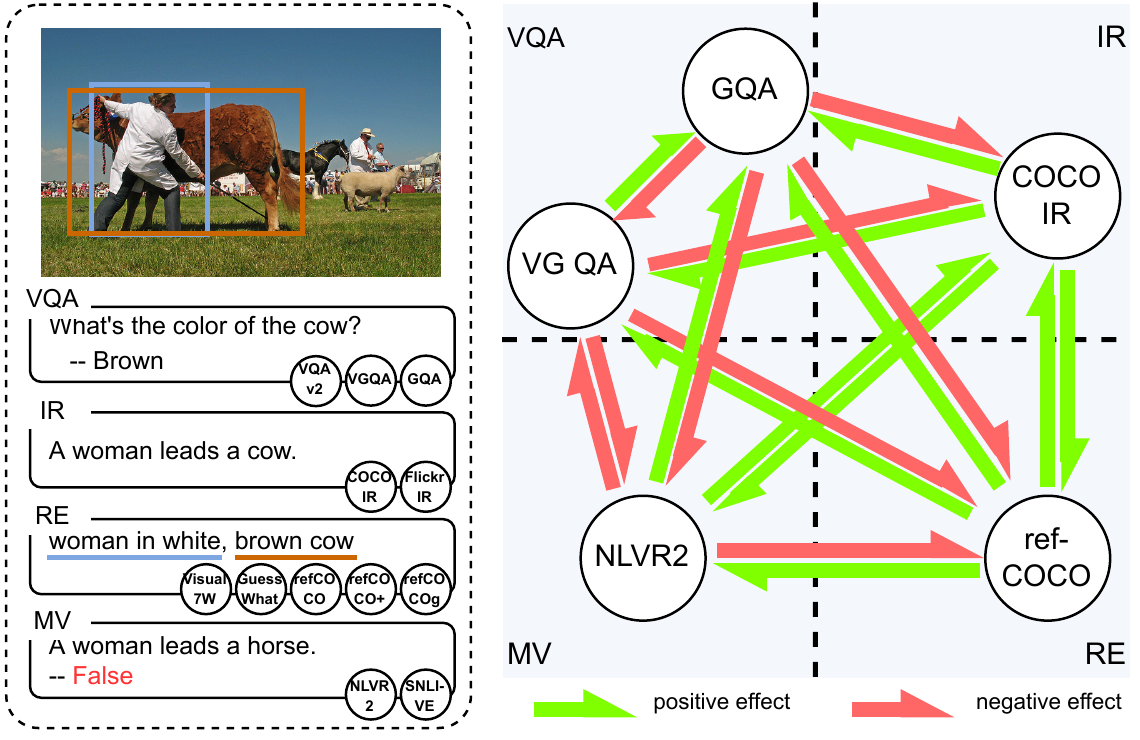}
  \caption{We explore the transferability among 12 vision-and-language tasks in 4 different groups: visual question answering (VQA), image retrieval (IR), referring expression (RE), and multi-modal verification (MV). Here, we illustrate the transferability among 5 tasks. Different tasks have different effect (positive or negative) on the other tasks.}
  \label{fig:intro}
\end{figure}

The problem with using hundreds of millions of samples for training is that the analysis, maintenance, processing, and especially understanding of the data is beyond human means. With rising concerns about large models encoding and perpetuating harmful representations towards historically discriminated groups \cite{bender2021dangers,prabhu2020large}, how data is handled acquires a crucial role. Knowing which data is being used, why, and for what means is now more important than ever.

We try to answer the question of \textit{whether more data is always better} by systematically analyzing the transferability within vision-and-language, which is the subset of tasks that require both visual and language understanding to be solved. For example, image captioning \cite{COCOIR} or visual question answering~\cite{VQA2.0}. In the last decade, dozens of high-quality vision-and-language datasets were collected, cleaned, and used as \textit{de facto} benchmarks for human-like reasoning \cite{ferraro-etal-2015-survey,mogadala2021trends}. Now, some of these datasets created with diverse motivations and purposes are coming together to train large vision-and-language models \cite{VilBERT,li2020oscar}. 

Whereas some tasks can improve their performance when a model is trained in a multi-dataset and multi-task protocol \cite{12in1}, to what extent and whether all vision-and-language tasks can benefit from this is still unclear. Our goal is to shed light on this question and explore the transferability of knowledge within vision-and-language tasks in a similar way as \cite{Taskonomy,Mensink2021FactorsOI} do for vision-only datasets. Specifically, we conduct hundreds of cross-experiments in which the performance of a target task trained under a dozen different initializations from different pre-trained source tasks is compared. 

Following \cite{12in1}, we divide vision-and-language tasks into 4 groups: visual question answering (VQA), image retrieval (IR), referring expression (RE), and multi-modal verification (MV), and we study both intra- and inter-group transferability. As shown in Figure \ref{fig:intro}, our results indicate that there is not yet a magic formula to consistently improve performance on all the datasets by transferring knowledge between tasks. In other words, while some target tasks benefit from a specific source task pre-training, others get harmed. Even within target tasks that are similar in terms of datasets and goals, different behavior is observed when the same source knowledge is transferred, and conversely, similar behaviors happen when different knowledge is transferred. This leads to the conclusion that more data is not always necessarily better for higher performance, as it depends on the training dataset's goal, nature, and size. 

From the experiments, we acquired several insights about the transferability of knowledge between visual-and-language models, which are summarized as follows:

\begin{itemize}[noitemsep,topsep=0pt]
    \item Tasks in the same group are more likely to help each other to improve performance. However, negative results  show that tasks with shared goals do not always contribute positively to one another. This indicates that having a shared goal is favorable, but not enough.
    \item In the inter-group experiments, we find that the RE tasks tend to have a positive effect on most of the tasks in other groups, while the MV group tends to receive a positive effect from other groups.
    \item While the best improvement is often given when knowledge is transferred within the same group, the worst results are concentrated on specific tasks, specially GQA~\cite{GQA}. We study why and how this happens.
    \item We detect that different random seeds strongly affect the numeric performance of each task, sometimes even more than the transfer learning itself. This urges to report vision-and-language results on multiple random configurations.
    \item We explore the effect of the data scale by down-sampling a large scale task. The results show increasing performance on all of the smaller scale tasks, which indicates that the dataset size is an important factor in knowledge transferability.
    \item Finally, we explore how different stages of training affect the performance of the target tasks. We discover that in some cases transferring knowledge at early stages of pre-training can bring benefits to the target task. When the model learns too much, the performance on the target task drops.
\end{itemize}

\section{Related work}

Knowledge transferability focuses on how a model that learns knowledge from source tasks can adapt to a new task. Existing research in this topic includes transfer learning \cite{FhdcDosovitskiy2021AnII,Kolesnikov2020BigT}, multi-task learning \cite{Strezoski2019ManyTL,Standley2020WhichTS} and meta-learning \cite{MAML,metaVQA}. Ideally, the more knowledge a model learns, the better performance it has. However, in practice, models are affected by several phenomena, such as catastrophic forgetting \cite{cataInNN,cataInNMT}, that limit their performance.
Our work is mainly related to the following two topics:

\paragraph{Transferability analysis.}
Transferability analysis studies how well the knowledge from a source task benefits a target task.
Zamir \etal~\cite{Taskonomy} proposed a method to analyze and utilize the transferability among 24 vision-only tasks on a single indoor scenes dataset. They pre-trained models in the source tasks, transferred them to the target tasks, and calculated the transferability by evaluating how well the model performed in the target task.
Following this idea, Mensink \etal~\cite{Mensink2021FactorsOI} studied the transferability between 20 real-world vision-only tasks. They analyzed three main factors: the image domain difference between source and target tasks, the task type, and the data size.

Whereas studies in \cite{Taskonomy,Mensink2021FactorsOI} were conducted on vision-only tasks, our goal is to explore multi-modality transferability in the vision-and-language domain. The particularity of multi-modal datasets is that knowledge needs to be transferred not only across tasks but also between modalities, which adds an extra layer of difficulty to the problem.

\begin{figure*}[ht]
\vspace{5pt}
    \centering
    \includegraphics[width=0.97\linewidth]{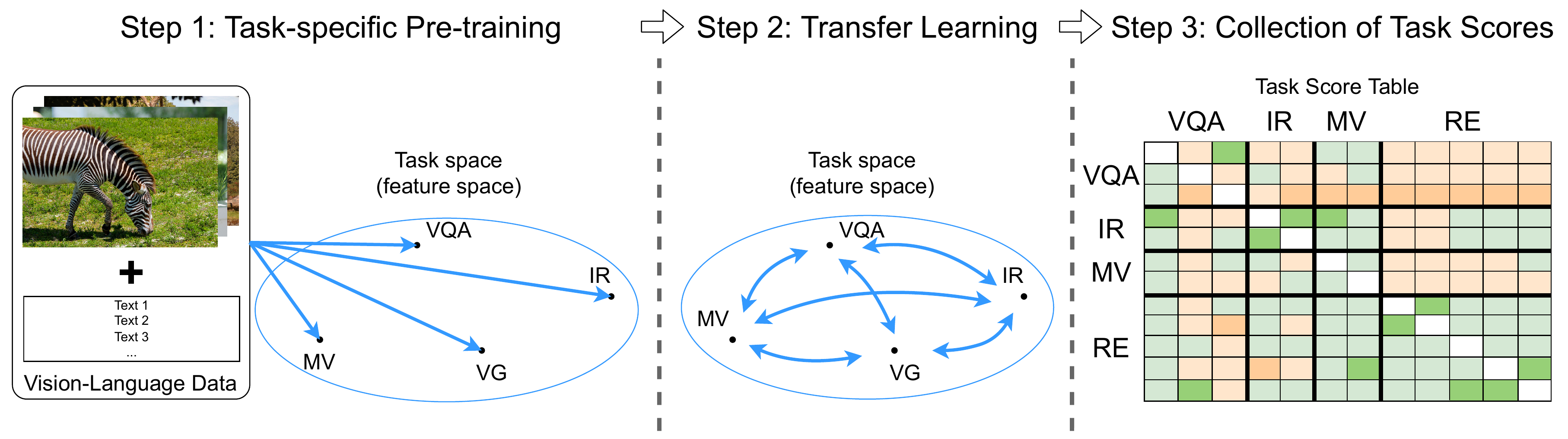}
    \vspace{5pt}
    \caption{Analysis of transferability relationships between tasks. In Step 1, we train 12 vision-and-language tasks independently. In Step 2, we use the models from Step 1 and fine-tune them on each of the other tasks. In Step 3, we form a transferability relation table for the 12 vision-and-language tasks in four groups: visual question answering (VQA), image retrieval (IR), multi-modal verification (MV) and referring expressions (RE).}
    \label{fig:workflow}
\end{figure*}

\paragraph{Vision-and-language tasks.}
Vision-and-language tasks require models to process both vision and language data. 
The most popular paradigm of knowledge transfer is to pre-train a model on a large dataset, and transfer it to a downstream task \cite{VilBERT,LXMERT,VisualBERT,VLBERT,UnicoderVL,UNITER,li2020oscar,Yu2021ERNIEViLKE,Zhang2021VinVLRV,VLAdapter}. For example, Lu \etal\cite{VilBERT} proposed a BERT-based vision-and-language model, and pre-trained it with three self-supervised tasks to learn knowledge from Google's Conceptual Captions dataset~\cite{GCC}. 
Following this work, many contributions were made in applying better text modeling \cite{Yu2021ERNIEViLKE}, better visual feature extraction \cite{Zhang2021VinVLRV}, and contrastive learning \cite{li2020oscar}.
Besides, there has been some work analyzing the knowledge transferability in specific tasks such as video question answering \cite{wu2021transferring}.

Our work is close to multi-task vision-and-language learning \cite{OmniNet,HDC,12in1,Hu2021UniTMM}.
Nguyen \etal~\cite{HDC} proposed a multi-task learning model with three vision-and-language tasks by choosing the best layers for each task.
In \cite{12in1}, a training strategy to prevent learning too much knowledge from converged tasks is proposed, resulting in a model trained on 12 vision-and-language tasks.
Following this idea, Hu \etal~\cite{Hu2021UniTMM} designed a unified transformer that can learn from either vision or text data. This model enables multi-task learning among vision-only, text-only, and vision-and-language tasks, and thus extends the knowledge that the vision-and-language model can learn.

None of the above work conducts a formal analysis on how the different tasks affect each other. In contrast, we thoroughly explore knowledge transferability among vision-and-language tasks and uncover insights that may be useful when applying knowledge transfer methods to vision-and-language.

\section{Methodology}

We study how the knowledge from a source task affects a target task. Formally, we define the problem as follows: 

Given a set $\mathcal{T}$ of vision-and-language tasks, we pick out a source task $s \in \mathcal{T}$ and a target task $t \in \mathcal{T}$.
We train a \textit{direct} model $m_t$ by training a model $m$ with the target task $t$. We also train a \textit{one-hop} model $m_{s\to t}$ by pre-training $m$ with $s$, and then with $t$. 
As shown in Figure~\ref{fig:workflow}, the performances of a pair of models $(m_t, m_{s \to t})$ are compared for all possible combinations of $s$ and $t$ in $\mathcal{T}$. Tasks are categorized in groups according to their main goal, so tasks with similar goals are assigned to the same group. %

\vspace{5pt}
\subsection{Group definitions}
\label{12 tasks}
Following \cite{12in1}, we study 12 vision-and-language tasks categorized in four groups.

\paragraph{Visual question answering group.} It contains tasks for answering questions based on visual data. Given an image and a related question, a model should select an answer from several candidates. 
The specific tasks used in this group are VQA v2 \cite{VQA2.0}, VG QA \cite{VGQA}, and GQA \cite{GQA}.

\paragraph{Image retrieval group.} It contains tasks for retrieving images according to a natural language sentence. Given a caption, the model needs to select the most representative image from a pool of images.
The tasks used in this group are COCO IR \cite{COCOIR} and Flickr30K IR \cite{FlickrIR}.

\paragraph{Referring expressions group.} It contains tasks for performing object detection based on a short sentence. Given a text and an image, the model should detect the corresponding region in the image described by the text.
The tasks used in this group are Visual 7w \cite{Visual7w}, RefCOCO(+/g) \cite{refCOCO1,refCOCO2} and GuessWhat \cite{GuessWhat}.

\paragraph{Multi-modal verification group.} It contains tasks for verifying the relationships between texts and images. Given one or more images and a referred text, the model should decide whether the text is correct or not.
The tasks used in this group are NLVR2 \cite{NLVR2} and SNLI-VE \cite{SNLIVE}.

\subsection{Model}
\label{Basic Model}
We follow the model structure in \cite{12in1}, consisting of a unified multi-modal encoder based on VilBERT \cite{VilBERT} with 12 different task-specific heads for corresponding tasks. The training goal is: %
\begin{equation}
    \arg\min_{\theta_e,\theta_{t}} L_{t}(\psi_{\theta_{t}}(\phi_{\theta_e}(V_t, S_t))),
\end{equation} %
where $V_t$ and $S_t$ are the image and text in the dataset of task $t$,  and $\theta_e$ and $\theta_{t}$ are the parameters of the encoder $\phi$ and the task $t$'s head $\psi$, respectively. $L_{t}$ is the loss of the task $t$.

\subsection{Workflow}
\label{Training strategy}
The workflow, as shown in Figure~\ref{fig:workflow}, is split into three steps: 1) task-specific pre-training, 2) transfer learning, and 3) collection of scores.

\vspace{-12pt}
\paragraph{Task-specific pre-training.} We pre-train each of the 12 tasks independently, \ie, each task $s \in \mathcal{T}$ is trained by its corresponding dataset and does not see any dataset from other tasks. We collect the trained models $m_{s}$ from each task as the pre-trained models, which learned task-specific knowledge from the source task. We also evaluate each \textit{direct} model $m_{t}$ as baselines for non-transferred knowledge.

\vspace{-12pt}
\paragraph{Transfer learning.} 
We fine-tune, again, each pre-trained model $m_s$. Given $m_s$ and the target task $t$, we get a final model $m_{s \to t}$ by fine-tuning $m_{s}$ with all of the training samples in task $t$.

\vspace{-12pt}
\paragraph{Collection of scores.} We categorize tasks into groups and evaluate all \textit{direct} models $m_t$ and \textit{one-hop} models $m_{s \to t}$ for all possible task pairs. Results are discussed in Section \ref{sec:results_transfer}. 

\begin{table}[]
    \centering
    \caption{Results of \textit{direct} model $m_t$ in the 12 tasks.}
    \renewcommand{\arraystretch}{1.2}
    \resizebox{!}{73pt}{
        \begin{tabular}{|cc|R{2cm}|R{1.7cm}|R{1.7cm}|}
            \hline
            \multicolumn{2}{|c|}{\textbf{10} different random seeds}                 & \multicolumn{1}{c|}{avg ± std} & \multicolumn{1}{c|}{max} & \multicolumn{1}{c|}{min}   \\ \hline
            \multicolumn{1}{|c|}{\multirow{12}{*}{Task}} & VQA v2 & \cellcolor[HTML]{E6F2F4}70.3±0.56                      & \cellcolor[HTML]{E6F2F4}70.71                    & \cellcolor[HTML]{E6F2F4}69.18 \\ \cline{2-2}
            \multicolumn{1}{|c|}{}                       & VG QA (Val)   & \cellcolor[HTML]{E6F2F4}33.5±0.48                      & \cellcolor[HTML]{E6F2F4}34.17                    & \cellcolor[HTML]{E6F2F4}32.86 \\ \cline{2-2}
            \multicolumn{1}{|c|}{}                       & GQA     & \cellcolor[HTML]{E6F2F4}58.1±0.53                      & \cellcolor[HTML]{E6F2F4}58.65                    & \cellcolor[HTML]{E6F2F4}57.10 \\ \cline{2-2}
            \multicolumn{1}{|c|}{}                       & COCO IR       & \cellcolor[HTML]{E6F2F4}90.4±0.77                      & \cellcolor[HTML]{E6F2F4}91.02                    & \cellcolor[HTML]{E6F2F4}89.12 \\ \cline{2-2}
            \multicolumn{1}{|c|}{}                       & Flickr30K IR  & \cellcolor[HTML]{E6F2F4}86.5±0.87                      & \cellcolor[HTML]{E6F2F4}87.24                    & \cellcolor[HTML]{E6F2F4}84.80 \\ \cline{2-2}
            \multicolumn{1}{|c|}{}                       & NLVR2         & \cellcolor[HTML]{E6F2F4}73.4±0.50                      & \cellcolor[HTML]{E6F2F4}74.11                    & \cellcolor[HTML]{E6F2F4}72.34 \\ \cline{2-2}
            \multicolumn{1}{|c|}{}                       & SNLI-VE       & \cellcolor[HTML]{E6F2F4}75.3±0.16                      & \cellcolor[HTML]{E6F2F4}75.64                    & \cellcolor[HTML]{E6F2F4}75.04 \\ \cline{2-2}
            \multicolumn{1}{|c|}{}                       & Visual7w      & \cellcolor[HTML]{E6F2F4}80.4±0.19                      & \cellcolor[HTML]{E6F2F4}80.63                    & \cellcolor[HTML]{E6F2F4}80.04 \\ \cline{2-2}
            \multicolumn{1}{|c|}{}                       & GuessWhat     & \cellcolor[HTML]{E6F2F4}62.3±0.17                      & \cellcolor[HTML]{E6F2F4}62.68                    & \cellcolor[HTML]{E6F2F4}62.14 \\ \cline{2-2}
            \multicolumn{1}{|c|}{}                       & refCOCO       & \cellcolor[HTML]{E6F2F4}77.7±0.30                      & \cellcolor[HTML]{E6F2F4}78.15                    & \cellcolor[HTML]{E6F2F4}77.13 \\ \cline{2-2}
            \multicolumn{1}{|c|}{}                       & refCOCO+      & \cellcolor[HTML]{E6F2F4}69.1±0.57                      & \cellcolor[HTML]{E6F2F4}69.68                    & \cellcolor[HTML]{E6F2F4}67.81 \\ \cline{2-2}
            \multicolumn{1}{|c|}{}                       & refCOCOg      & \cellcolor[HTML]{E6F2F4}71.6±0.63                      & \cellcolor[HTML]{E6F2F4}72.50                    & \cellcolor[HTML]{E6F2F4}70.50 \\ \hline
        \end{tabular}
    }
    \label{table:result_10seeds}
\end{table}

\section{Experiments}

\paragraph{Datasets.}
We use the same set of datasets as \cite{12in1}, including the training and test sets of the 12 tasks. To prevent leaking data from the test set into the training set, the overlapping samples from the different tasks were removed from the training sets. Note that the original test sets were not changed during this cleaning process. 
For training and validation sets, VQA v2, VG QA, COCO IR, and NLVR2 have about 100,000 images; GQA and GuessWhat have about 60,000 images; Flickr30K IR and SNLI-VE have about 30,000 images; and refCOCO, refCOCO+. refCOCOg and Visual7w have about 15,000 images. More details about each dataset can be found in the supplementary material.
More details are shown in the Appendix.

\paragraph{Experimental settings.}
We follow most of the settings in~\cite{12in1}. We modify the batch size to $1/4$ to fit the training in our server.\footnote{We use a single server with 4 16GB NVIDIA P100 GPUs.} Pre-trained models $m_s$ are trained for 6 epochs, which is enough for convergence.
To ensure models $m_{s \to t}$ learn task-specific knowledge well, we use the models with the best performance in the validation set, except for VG QA that is evaluated on the validation set, and thus the model at the 6th epoch is used. All of the models are seen converged in their corresponding tasks.
We train every model with three different random seeds and report results by their mean and standard deviation.

\paragraph{Evaluation metrics.}
We use accuracy for tasks in the VQA group and the MV group. For the IR group, we use Recall@5. For RE group, we follow \cite{12in1,refCOCO1,refCOCO2} and compute the score based on intersection over union (IOU) between the ground truth and the prediction. 

\subsection{Random seed}
\label{sec:results_stability}

\begin{figure*}[ht]
    \centering
    \includegraphics[width=0.95\textwidth]{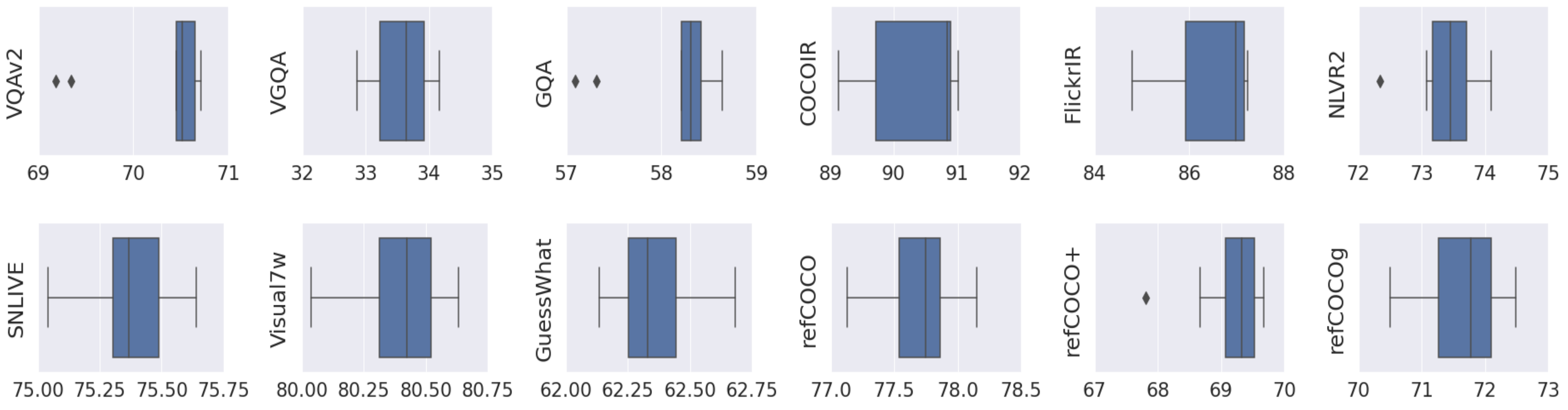}
    \caption{Box plots of the 12 tasks trained with 10 random seeds showing a big gap between the best and the worst scores.}
    \label{fig:result_10seeds}
\end{figure*}

Preliminary results showed large variations in performance when models are trained under different random initializations, as also shown in \cite{RandomSEEDs}.
Thus, before proceeding with the transferrability experiments, we first explore the instability of vision-and-language tasks and their sensibility to random seeds. We train each direct model, $m_t$ for all $t \in \mathcal{T}$, 10 times with different random seeds. 
The results are shown in Figure~\ref{fig:result_10seeds}. Although most of the tasks present a gap larger than $1\%$ between the maximum and the minimum score, most of the scores in each task are concentrated in a small region. 
More details are shown in the box plots in Table~\ref{table:result_10seeds}. There are nine tasks that have a gap larger than $1\%$. Among them, Flickr30K IR is the one that fluctuates the most with a gap of $2.44\%$ and a standard deviation of $0.87$. This reveals that experiments on a single run may not be reliable to extract conclusions about model performance.
In general, we found that the random seed has a big impact on the evaluation of vision-and-language task. To ensure our results are reliable, we run each experiment three times.

\subsection{Results by group}
\label{sec:results_transfer}

For the transferability experiments, we collect results from 12 \textit{direct} models $m_t$ and 132 \textit{one-hop} models $m_{s \to t}$ and present them  in Tables \ref{table:result_vqa}, \ref{table:result_ir}, \ref{table:result_mv}, and \ref{table:result_re}. We use the results of $m_t$ (Row ``\textit{direct} model $m_t$") as the baseline. We rely on a color scheme to illustrate the comparative performance of the transferred models $m_{s \to t}$: deep green for the best scores of each column, \ie, the best results of each task in transfer learning, and deep red for the worst results. For the rest of entries of the tables, light green and light red indicate better and worse performance than the baseline.

\paragraph{Visual question answering group.}
Table \ref{table:result_vqa} shows the results in the VQA group. 
VQA v2 and GQA benefit each other, but they do not improve the VG QA performance. In fact, GQA has the worst effect on VG QA among the 12 tasks. VG QA gets its best performance with the help of refCOCOg, indicating that even though it is commonly seen as a VQA task, it may be closer to the RE group. 
When tasks in the VQA group are the target task, the source tasks have a consistent effect on each of them, \eg COCO IR give the best effect to VQA v2, while give negative effect to both VG QA and GQA. In contrast, GuessWhat give the worst effect to GQA while give positive effect to VQA v2.
More specifically, VQA v2 and VG QA show contrary behavior: VQA v2 gets positive effect from all of the tasks outside the VQA group, while only refCOCO and refCOCO+ give VG QA positive effect.
This indicates that although VQA, VG QA, and GQA have the same type of training goal, 
their underlying knowledge may be very different, and thus receiving different contributions from the same source task. 
Finally, even though tasks in the VQA group are the largest in terms of training samples, when they act as the source task, they tend to have a negative impact on the other group tasks (Tables \ref{table:result_ir}, \ref{table:result_mv}, and \ref{table:result_re}), indicating that large training sets are not a guarantee for better transfer.

\begin{table*}[t]
    \caption{Knowledge transferability results per group. Results of $m_{row\  \to \  column}$ (row 2-13, green or red color) are compared with the \textit{direct} model $m_{column}$ (row 1, blue color) and assigned green (when the average score is higher than $m_{column}$) or red (when the average score is lower than $m_{column}$). Deep green/red shows the best/worst score in each column.}
    \label{tabel:12 results}
    \begin{subtable}[h]{0.5\textwidth}
        \centering
        \caption{Visual question answering group}
        \renewcommand{\arraystretch}{1.2}
        \resizebox{!}{73pt}{
            \begin{tabular}{|cc|R{2.5cm}|R{2.5cm}|R{2.5cm}|}
                \hline
                \multicolumn{2}{|c|}{}                                                                                                             & \multicolumn{3}{c|}{Target Task $t$}  \\ \cline{3-5} 
                \multicolumn{2}{|c|}{\multirow{-2}{*}{\cellcolor[HTML]{FFFFFF}{\color[HTML]{333333} \textbf{avg ± std}}}}                 & \multicolumn{1}{c|}{VQA v2}                            & \multicolumn{1}{c|}{VG QA (Val)}                       & \multicolumn{1}{c|}{GQA}                               \\ \hline
                \multicolumn{2}{|c|}{\textit{direct} model $m_t$}                                                                              & \multicolumn{1}{r|}{\cellcolor[HTML]{E6F2F4}69.6±0.71} & \multicolumn{1}{r|}{\cellcolor[HTML]{E6F2F4}33.7±0.74} & \cellcolor[HTML]{E6F2F4}57.5±0.59 \\ \hline
                \multicolumn{1}{|c|}{}                                                                            & VQA v2      & -                                                      & \cellcolor[HTML]{FCE5CD}33.5±0.55                      & \cellcolor[HTML]{8AC677}58.2±0.21 \\ \cline{2-2}
                \multicolumn{1}{|c|}{}                                                                            & VG QA        & \cellcolor[HTML]{D9EAD3}70.0±0.33                      & -                                                      & \cellcolor[HTML]{FCE5CD}57.5±0.57 \\ \cline{2-2}
                \multicolumn{1}{|c|}{}                                                                            & GQA          & \cellcolor[HTML]{D9EAD3}69.7±0.16                      & \cellcolor[HTML]{F9CB9C}33.0±0.76                      & -                                 \\ \cline{2-5} 
                \multicolumn{1}{|c|}{}                                                                            & COCO IR      & \cellcolor[HTML]{8AC677}70.5±0.56                      & \cellcolor[HTML]{FCE5CD}33.6±0.52                      & \cellcolor[HTML]{FCE5CD}57.4±0.49 \\ \cline{2-2}
                \multicolumn{1}{|c|}{}                                                                            & Flickr30K IR & \cellcolor[HTML]{D9EAD3}70.3±0.32                      & \cellcolor[HTML]{FCE5CD}33.4±0.69                      & \cellcolor[HTML]{D9EAD3}57.6±0.35 \\ \cline{2-5} 
                \multicolumn{1}{|c|}{}                                                                            & NLVR2        & \cellcolor[HTML]{D9EAD3}69.9±0.34                      & \cellcolor[HTML]{FCE5CD}33.4±0.35                      & \cellcolor[HTML]{D9EAD3}57.5±0.43 \\ \cline{2-2}
                \multicolumn{1}{|c|}{}                                                                            & SNLI-VE      & \cellcolor[HTML]{D9EAD3}69.9±0.68                      & \cellcolor[HTML]{FCE5CD}33.3±0.51                      & \cellcolor[HTML]{FCE5CD}57.3±0.32 \\ \cline{2-5} 
                \multicolumn{1}{|c|}{}                                                                            & Visual7w     & \cellcolor[HTML]{D9EAD3}70.2±0.25                      & \cellcolor[HTML]{FCE5CD}33.5±0.94                      & \cellcolor[HTML]{D9EAD3}57.7±0.57 \\ \cline{2-2}
                \multicolumn{1}{|c|}{}                                                                            & GuessWhat    & \cellcolor[HTML]{D9EAD3}69.7±0.61                      & \cellcolor[HTML]{FCE5CD}33.5±0.06                      & \cellcolor[HTML]{F9CB9C}56.9±0.33 \\ \cline{2-2}
                \multicolumn{1}{|c|}{}                                                                            & refCOCO      & \cellcolor[HTML]{D9EAD3}70.2±0.22                      & \cellcolor[HTML]{D9EAD3}33.7±0.29                      & \cellcolor[HTML]{FCE5CD}57.4±0.21 \\ \cline{2-2}
                \multicolumn{1}{|c|}{}                                                                            & refCOCO+     & \cellcolor[HTML]{D9EAD3}70.1±0.44                      & \cellcolor[HTML]{FCE5CD}33.3±0.05                      & \cellcolor[HTML]{FCE5CD}57.2±0.41 \\ \cline{2-2}
                \multicolumn{1}{|c|}{\multirow{-12}{*}{\begin{tabular}[c]{@{}c@{}}Source \\ task s\end{tabular}}} & refCOCOg     & \cellcolor[HTML]{D9EAD3}69.7±0.30                      & \cellcolor[HTML]{8AC677}34.0±0.46                      & \cellcolor[HTML]{FCE5CD}57.4±1.07 \\ \hline
            \end{tabular}
        }
        \label{table:result_vqa}
    \end{subtable}
    \hfill
    \begin{subtable}[h]{0.45\textwidth}
        \centering
        \caption{Image retrieval group}
        \renewcommand{\arraystretch}{1.2}
        \resizebox{!}{73pt}{
            \begin{tabular}{|cc|R{2.5cm}|R{2.5cm}|}
            \hline
            \multicolumn{2}{|c|}{}                                                                                             & \multicolumn{2}{c|}{Target task $t$}                                                         \\ \cline{3-4} 
            \multicolumn{2}{|c|}{\multirow{-2}{*}{\begin{tabular}[c]{@{}c@{}}\textbf{avg ± std}\end{tabular}}} & \multicolumn{1}{c|}{COCO IR}                           & \multicolumn{1}{c|}{Flickr30K IR} \\ \hline
            \multicolumn{2}{|c|}{\textit{direct} model $m_t$}                                                                        & \multicolumn{1}{r|}{\cellcolor[HTML]{E6F2F4}89.2±0.16} & \cellcolor[HTML]{E6F2F4}85.3±0.48 \\ \hline
            \multicolumn{1}{|c|}{}                                                                             & VQA v2       & \cellcolor[HTML]{FCE5CD}89.0±1.43                      & \cellcolor[HTML]{FCE5CD}84.8±1.04 \\ \cline{2-2}
            \multicolumn{1}{|c|}{}                                                                             & VG QA         & \cellcolor[HTML]{D9EAD3}90.0±0.11                      & \cellcolor[HTML]{FCE5CD}84.3±0.77 \\ \cline{2-2}
            \multicolumn{1}{|c|}{}                                                                             & GQA           & \cellcolor[HTML]{FCE5CD}88.9±1.17                      & \cellcolor[HTML]{F9CB9C}82.9±1.46 \\ \cline{2-4} 
            \multicolumn{1}{|c|}{}                                                                             & COCO IR       & -                                                      & \cellcolor[HTML]{8AC677}86.8±1.56 \\ \cline{2-2}
            \multicolumn{1}{|c|}{}                                                                             & Flickr30K IR  & \cellcolor[HTML]{8AC677}90.1±1.30                      & -                                 \\ \cline{2-4} 
            \multicolumn{1}{|c|}{}                                                                             & NLVR2         & \cellcolor[HTML]{D9EAD3}89.7±1.16                      & \cellcolor[HTML]{FCE5CD}84.8±1.09 \\ \cline{2-2}
            \multicolumn{1}{|c|}{}                                                                             & SNLI-VE       & \cellcolor[HTML]{FCE5CD}89.2±0.51                      & \cellcolor[HTML]{D9EAD3}85.5±1.80 \\ \cline{2-4} 
            \multicolumn{1}{|c|}{}                                                                             & Visual7w      & \cellcolor[HTML]{D9EAD3}89.8±0.45                      & \cellcolor[HTML]{D9EAD3}85.6±1.06 \\ \cline{2-2}
            \multicolumn{1}{|c|}{}                                                                             & GuessWhat     & \cellcolor[HTML]{D9EAD3}89.5±0.45                      & \cellcolor[HTML]{FCE5CD}85.1±2.09 \\ \cline{2-2}
            \multicolumn{1}{|c|}{}                                                                             & refCOCO       & \cellcolor[HTML]{D9EAD3}90.1±0.83                      & \cellcolor[HTML]{D9EAD3}85.4±1.33 \\ \cline{2-2}
            \multicolumn{1}{|c|}{}                                                                             & refCOCO+      & \cellcolor[HTML]{F9CB9C}88.8±1.93                      & \cellcolor[HTML]{FCE5CD}84.9±2.27 \\ \cline{2-2}
            \multicolumn{1}{|c|}{\multirow{-12}{*}{\begin{tabular}[c]{@{}c@{}}Source \\ task $s$\end{tabular}}}  & refCOCOg      & \cellcolor[HTML]{D9EAD3}89.1±2.19                      & \cellcolor[HTML]{D9EAD3}84.9±1.24 \\ \hline
            \end{tabular}
        }
        \label{table:result_ir}
    \end{subtable}
    \hfill
    \\[0.5cm]
    \begin{subtable}[h]{0.40\textwidth}
        \centering
        \caption{Multi-modal verification group}
        \renewcommand{\arraystretch}{1.2}
        \resizebox{!}{73pt}{
            \begin{tabular}{|c|c|R{2.5cm}|R{2.5cm}|}
            \hline
            \multicolumn{2}{|c|}{}                                                                                             & \multicolumn{2}{c|}{Target task $t$}                                                         \\ \cline{3-4} 
            \multicolumn{2}{|c|}{\multirow{-2}{*}{\begin{tabular}[c]{@{}c@{}}\textbf{avg ± std}\end{tabular}}} & \multicolumn{1}{c|}{NLVR2}                             & \multicolumn{1}{c|}{SNLI-VE}      \\ \hline
            \multicolumn{2}{|c|}{\textit{direct} model $m_t$}                                                                        & \multicolumn{1}{r|}{\cellcolor[HTML]{E6F2F4}72.8±0.48} & \cellcolor[HTML]{E6F2F4}75.3±0.11 \\ \hline
            \multicolumn{1}{|c|}{}                                                                             & VQA v2       & \cellcolor[HTML]{D9EAD3}73.7±0.60                      & \cellcolor[HTML]{D9EAD3}75.4±0.23 \\ \cline{2-2}
            \multicolumn{1}{|c|}{}                                                                             & VG QA         & \cellcolor[HTML]{FCE5CD}72.1±0.69                      & \cellcolor[HTML]{D9EAD3}75.7±0.05 \\ \cline{2-2}
            \multicolumn{1}{|c|}{}                                                                             & GQA           & \cellcolor[HTML]{F9CB9C}72.1±0.76                      & \cellcolor[HTML]{F9CB9C}75.3±0.67 \\ \cline{2-4} 
            \multicolumn{1}{|c|}{}                                                                             & COCO IR       & \cellcolor[HTML]{8AC677}75.3±0.24                      & \cellcolor[HTML]{D9EAD3}76.1±0.11 \\ \cline{2-2}
            \multicolumn{1}{|c|}{}                                                                             & Flickr30K IR  & \cellcolor[HTML]{D9EAD3}74.0±0.65                      & \cellcolor[HTML]{D9EAD3}75.8±0.22 \\ \cline{2-4} 
            \multicolumn{1}{|c|}{}                                                                             & NLVR2         & -                                                      & \cellcolor[HTML]{D9EAD3}75.9±0.06 \\ \cline{2-2}
            \multicolumn{1}{|c|}{}                                                                             & SNLI-VE       & \cellcolor[HTML]{D9EAD3}73.9±0.24                      & -                                 \\ \cline{2-4} 
            \multicolumn{1}{|c|}{}                                                                             & Visual7w      & \cellcolor[HTML]{D9EAD3}73.9±0.71                      & \cellcolor[HTML]{D9EAD3}76.1±0.26 \\ \cline{2-2}
            \multicolumn{1}{|c|}{}                                                                             & GuessWhat     & \cellcolor[HTML]{D9EAD3}73.2±0.31                      & \cellcolor[HTML]{D9EAD3}75.9±0.34 \\ \cline{2-2}
            \multicolumn{1}{|c|}{}                                                                             & refCOCO       & \cellcolor[HTML]{D9EAD3}73.7±0.28                      & \cellcolor[HTML]{D9EAD3}76.0±0.33 \\ \cline{2-2}
            \multicolumn{1}{|c|}{}                                                                             & refCOCO+      & \cellcolor[HTML]{D9EAD3}74.4±0.22                      & \cellcolor[HTML]{8AC677}76.1±0.14 \\ \cline{2-2}
            \multicolumn{1}{|c|}{\multirow{-12}{*}{\begin{tabular}[c]{@{}c@{}}Source \\ task $s$\end{tabular}}}  & refCOCOg      & \cellcolor[HTML]{D9EAD3}74.1±0.59                      & \cellcolor[HTML]{D9EAD3}75.7±0.20 \\ \hline
            \end{tabular}}
        \label{table:result_mv}
    \end{subtable}
    \hfill
    \begin{subtable}[h]{0.55\textwidth}
        \centering
        \caption{Referring expression group}
        \renewcommand{\arraystretch}{1.2}
        \resizebox{!}{73pt}{
            \begin{tabular}{|c|c|r|r|r|r|r|}
            \hline
                \multicolumn{2}{|c|}{}                                                                                             & \multicolumn{5}{c|}{Target task $t$}                                                                                                                                                                                                                                    \\ \cline{3-7} 
                \multicolumn{2}{|c|}{\multirow{-2}{*}{\begin{tabular}[c]{@{}c@{}}\textbf{avg ± std}\end{tabular}}} & \multicolumn{1}{c|}{Visual7w}                          & \multicolumn{1}{c|}{GuessWhat}                         & \multicolumn{1}{c|}{refCOCO}                           & \multicolumn{1}{c|}{refCOCO+}                          & \multicolumn{1}{c|}{refCOCOg}     \\ \hline
                \multicolumn{2}{|c|}{\textit{direct} model $m_t$}                                                                        & \multicolumn{1}{r|}{\cellcolor[HTML]{E6F2F4}80.1±0.13} & \multicolumn{1}{r|}{\cellcolor[HTML]{E6F2F4}62.3±0.08} & \multicolumn{1}{r|}{\cellcolor[HTML]{E6F2F4}77.3±0.19} & \multicolumn{1}{r|}{\cellcolor[HTML]{E6F2F4}68.5±0.72} & \cellcolor[HTML]{E6F2F4}70.8±0.35 \\ \hline
                \multicolumn{1}{|c|}{}                                                                             & VQA v2       & \cellcolor[HTML]{FCE5CD}79.5±0.51                      & \cellcolor[HTML]{FCE5CD}60.8±0.62                      & \cellcolor[HTML]{FCE5CD}76.8±0.62                      & \cellcolor[HTML]{FCE5CD}67.7±1.12                      & \cellcolor[HTML]{FCE5CD}70.5±0.12 \\ \cline{2-2}
                \multicolumn{1}{|c|}{}                                                                             & VG QA         & \cellcolor[HTML]{FCE5CD}79.9±0.91                      & \cellcolor[HTML]{FCE5CD}60.9±0.66                      & \cellcolor[HTML]{FCE5CD}76.9±0.57                      & \cellcolor[HTML]{FCE5CD}68.0±0.81                      & \cellcolor[HTML]{FCE5CD}70.8±0.33 \\ \cline{2-2}
                \multicolumn{1}{|c|}{}                                                                             & GQA           & \cellcolor[HTML]{F9CB9C}79.0±0.12                      & \cellcolor[HTML]{F9CB9C}60.7±0.26                      & \cellcolor[HTML]{F9CB9C}76.7±0.32                      & \cellcolor[HTML]{F9CB9C}67.2±0.22                      & \cellcolor[HTML]{F9CB9C}70.0±0.42 \\ \cline{2-7} 
                \multicolumn{1}{|c|}{}                                                                             & COCO IR       & \cellcolor[HTML]{FCE5CD}79.5±0.34                      & \cellcolor[HTML]{FCE5CD}62.0±0.36                      & \cellcolor[HTML]{D9EAD3}77.4±0.59                      & \cellcolor[HTML]{D9EAD3}69.4±0.33                      & \cellcolor[HTML]{D9EAD3}72.0±0.25 \\ \cline{2-2}
                \multicolumn{1}{|c|}{}                                                                             & Flickr30K IR  & \cellcolor[HTML]{FCE5CD}79.8±0.17                      & \cellcolor[HTML]{FCE5CD}62.3±0.16                      & \cellcolor[HTML]{D9EAD3}77.3±0.18                      & \cellcolor[HTML]{D9EAD3}68.7±0.22                      & \cellcolor[HTML]{D9EAD3}71.3±0.23 \\ \cline{2-7} 
                \multicolumn{1}{|c|}{}                                                                             & NLVR2         & \cellcolor[HTML]{FCE5CD}79.4±0.27                      & \cellcolor[HTML]{FCE5CD}62.0±0.17                      & \cellcolor[HTML]{FCE5CD}77.1±0.40                      & \cellcolor[HTML]{FCE5CD}68.4±0.40                      & \cellcolor[HTML]{D9EAD3}70.9±0.10 \\ \cline{2-2}
                \multicolumn{1}{|c|}{}                                                                             & SNLI-VE       & \cellcolor[HTML]{FCE5CD}79.2±0.60                      & \cellcolor[HTML]{FCE5CD}61.2±0.40                      & \cellcolor[HTML]{FCE5CD}76.8±0.43                      & \cellcolor[HTML]{FCE5CD}67.2±0.41                      & \cellcolor[HTML]{FCE5CD}70.4±0.65 \\ \cline{2-7} 
                \multicolumn{1}{|c|}{}                                                                             & Visual7w      & -                                                      & \cellcolor[HTML]{8AC677}63.0±0.36                      & \cellcolor[HTML]{D9EAD3}78.1±0.39                      & \cellcolor[HTML]{D9EAD3}69.4±0.06                      & \cellcolor[HTML]{D9EAD3}72.8±0.30 \\ \cline{2-2}
                \multicolumn{1}{|c|}{}                                                                             & GuessWhat     & \cellcolor[HTML]{8AC677}80.8±0.05                      & -                                                      & \cellcolor[HTML]{D9EAD3}78.1±0.13                      & \cellcolor[HTML]{D9EAD3}69.1±0.13                      & \cellcolor[HTML]{D9EAD3}72.2±0.06 \\ \cline{2-2}
                \multicolumn{1}{|c|}{}                                                                             & refCOCO       & \cellcolor[HTML]{D9EAD3}80.3±0.03                      & \cellcolor[HTML]{D9EAD3}62.6±0.29                      & -                                                      & \cellcolor[HTML]{D9EAD3}69.5±0.23                      & \cellcolor[HTML]{D9EAD3}72.4±0.29 \\ \cline{2-2}
                \multicolumn{1}{|c|}{}                                                                             & refCOCO+      & \cellcolor[HTML]{D9EAD3}80.4±0.17                      & \cellcolor[HTML]{D9EAD3}62.5±0.29                      & \cellcolor[HTML]{D9EAD3}77.8±0.35                      & -                                                      & \cellcolor[HTML]{8AC677}73.0±0.19 \\ \cline{2-2}
                \multicolumn{1}{|c|}{\multirow{-12}{*}{\begin{tabular}[c]{@{}c@{}}Source \\ task $s$\end{tabular}}}  & refCOCOg      & \cellcolor[HTML]{D9EAD3}80.5±0.35                      & \cellcolor[HTML]{D9EAD3}62.8±0.15                      & \cellcolor[HTML]{8AC677}78.3±0.14                      & \cellcolor[HTML]{8AC677}69.7±0.35                      & -                                 \\ \hline
            \end{tabular}}
        \label{table:result_re}
    \end{subtable}
\end{table*}

\paragraph{Image retrieval group.}
Table \ref{table:result_ir} summarizes the performance of the IR group. Both tasks in this group help each other. Also, as source tasks, they show similar behavior, with a tendency to improve other tasks.
However, the results in this group show the largest variance. 
On one hand, as target tasks, only the VQA group has a consistent negative impact on Flickr30K IR. On the other hand, the standard deviation scores in COCO IR and Flicker30K IR are usually larger than in other groups. The standard deviation scores of $m_{\text{COCO IR} \to \text{Flickr30K IR}}$ and $m_{\text{Flickr30K IR} \to \text{COCO IR}}$ tend to be larger than tasks in other groups. 

\paragraph{Multi-modal verification group.}
Table \ref{table:result_mv} shows the performance of the MV group. Except for GQA, most of the source tasks have a positive effect on the two MV tasks. NLVR2 and SNLI-VE also improve each other, but the effect is not larger than the ones from COCO IR and refCOCO+. This may be in part because NLVR2 and SNLI-VE are considerably different: NLVR2 is a binary classification task that verifies if a comment describes a fact among multiple images, while SNLI-VE is a ternary classification task that verifies how well a comment describes an image. Another reason may be due to the data distributions: NLVR2's images are from ILSVRC 2014~\cite{ILSVRC2014}, while SNLI-VE's are from Flickr30K~\cite{Flickr30K}.

\paragraph{Referring expressions group.}
Table \ref{table:result_re} shows the scores of the RE group. All the tasks in this group benefit from transferred knowledge in the same group. The improvements within this group, especially among refCOCO, refCOCO+, and refCOCOg, are larger than those from tasks in other groups. %
However, all tasks in the VQA and the MV groups have a negative effect on the RE group (except $m_{\text{NLVR2} \to \text{refCOCOg}}$). RE tasks also receive the worst effect from the GQA task. 
Tasks in this group usually have a positive outcome on the tasks in other groups, according to Table \ref{table:result_vqa}, \ref{table:result_ir}, and \ref{table:result_mv}.
This may be because of the nature of the group: RE tasks aim to find image regions given a text, which can be helpful to VQA, IR, and MV.

\subsection{Main observations}
\label{sec:observation}

\begin{description}
\item[Observation 1.] \textbf{Intra-group analysis: tasks in the same group tend to improve each other, but not always.}
Tasks in the IR, MV, and RE groups help other tasks in the same group to get a better performance. Tasks in the VQA group, however, show different behavior: only half of the intra-class relationships are positive. This indicates that: 1) the defined task groups based on shared goals may be superficial and not a good representation of the internal type of knowledge in each task, and 2) having a shared goal may be favorable, but it is not enough for successfully transferring knowledge between tasks.

\item[Observation 2.] \textbf{Inter-group analysis: some groups are more prone to help, while others make disservice.}
For example, while tasks in the RE group usually give positive effect to most of the tasks that are in other groups, tasks in the VQA group produce no benefit to the tasks in the RE group, and only one task in the MV group (NLVR2) give slightly positive effect to a task in the RE group  (refCOCOg). This indicates that the knowledge in certain groups, such as RE, may be more general, and thus easier to transfer, than task-specific knowledge from other, \eg VQA, groups.

\item[Observation 3.] \textbf{Benefits in knowledge transferability are no reciprocal.} 
For example, VQA v2 receives positive effect from all of the other 11 tasks, but it contributes negatively to most of these tasks, except GQA, NLVR2 and SNLI-VE. 
The same happens between the MV and RE groups. RE consistently improves the MV group, including the best effect on SNLI-VE from refCOCO+. However, the MV group harms all the tasks in the RE group, except $m_{\text{NLVR2} \to \text{refCOCOg}}$. This is consistent with the observations in \cite{12in1}.

\item[Observation 4.] \textbf{The best effect tends to come within the group, while the worst effect is usually from GQA.}
The best results for each task usually are from a source task in the same group, which reinforces the idea that tasks with the same target tend to benefit each other more (Observation 1). 
The worst results, however, are usually caused by GQA. Many reasons may cause this, such as the difference in the data scale between GQA and the rest of the tasks, or that the knowledge for solving GQA may be too specific.
To better understand phenomena, we conduct additional experiments in Section~\ref{sec:results_data_scale} and Section~\ref{sec:results_epoch}.
\end{description}

\subsection{Data scale}
\label{sec:results_data_scale}
\begin{figure}[t]
    \centering
    \includegraphics[width=0.9\linewidth]{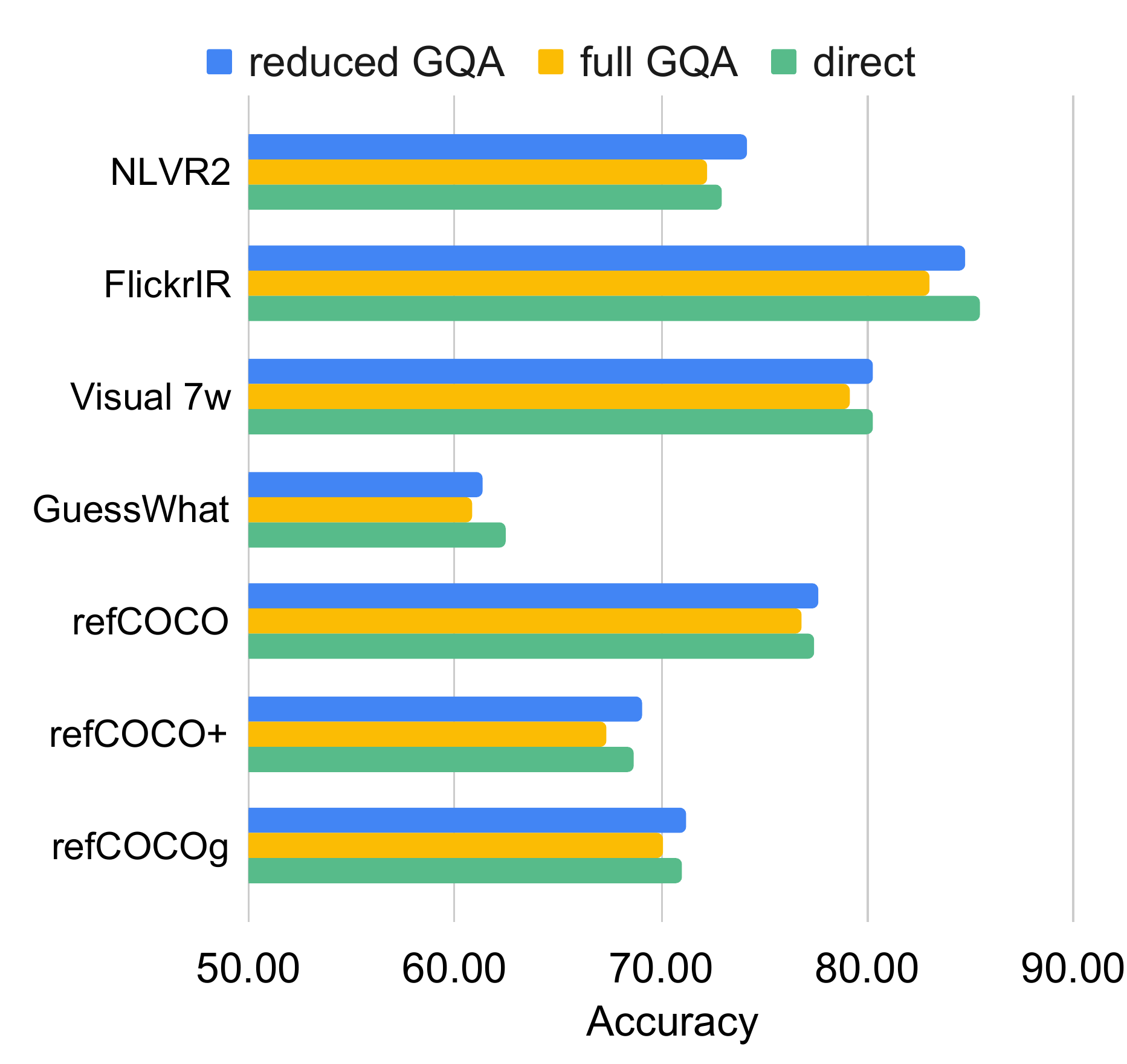}
    \caption{Accuracy of seven tasks pre-trained with a smaller set of GQA (reduced GQA), the full set of GQA (full GQA), and without pre-training (direct).}
    \label{fig:result_reduceGQA}
\end{figure}

Next, we investigate the effect of data scale on the transferability of knowledge. 
As discussed in Section~\ref{sec:observation}, GQA pre-training tends to harm the most the rest of the tasks. We speculate that one of the reasons may be because GQA has a much larger training set than the other tasks. To investigate this hypothesis, we use GQA as the source task and down sample its training set from $962,928$ to $96,221$, which is close to the scale of seven tasks: refCOCO, refCOCO+, refCOCOg, Visual7W, GuessWhat, NLVR2, and Flickr30K IR. We pre-train models with the reduced and full GQA training sets. 
The full GQA model and the reduced GQA model are then trained again on the seven tasks above in the same way as in Section \ref{Training strategy}. We also compare them against their the direct models. 

Figure \ref{fig:result_reduceGQA} shows the accuracy of these seven tasks pre-trained on the reduced GQA model, the full GQA model, and the direct models. All models pre-trained on the reduced GQA get better performance than those pre-trained on the full GQA, which indicates that data scale is an import factor in the transferability. When comparing the models derived from the reduced GQA with the direct models, the reduced models improve the performance for four of seven tasks (NLVR2, refCOCO, refCOCO+, and refCOCOg), showing that GQA can also contribute positively as a source task.

\subsection{Training epoch}
\label{sec:results_epoch}
\begin{figure}[t]
    \centering
    \includegraphics[width=0.95\linewidth]{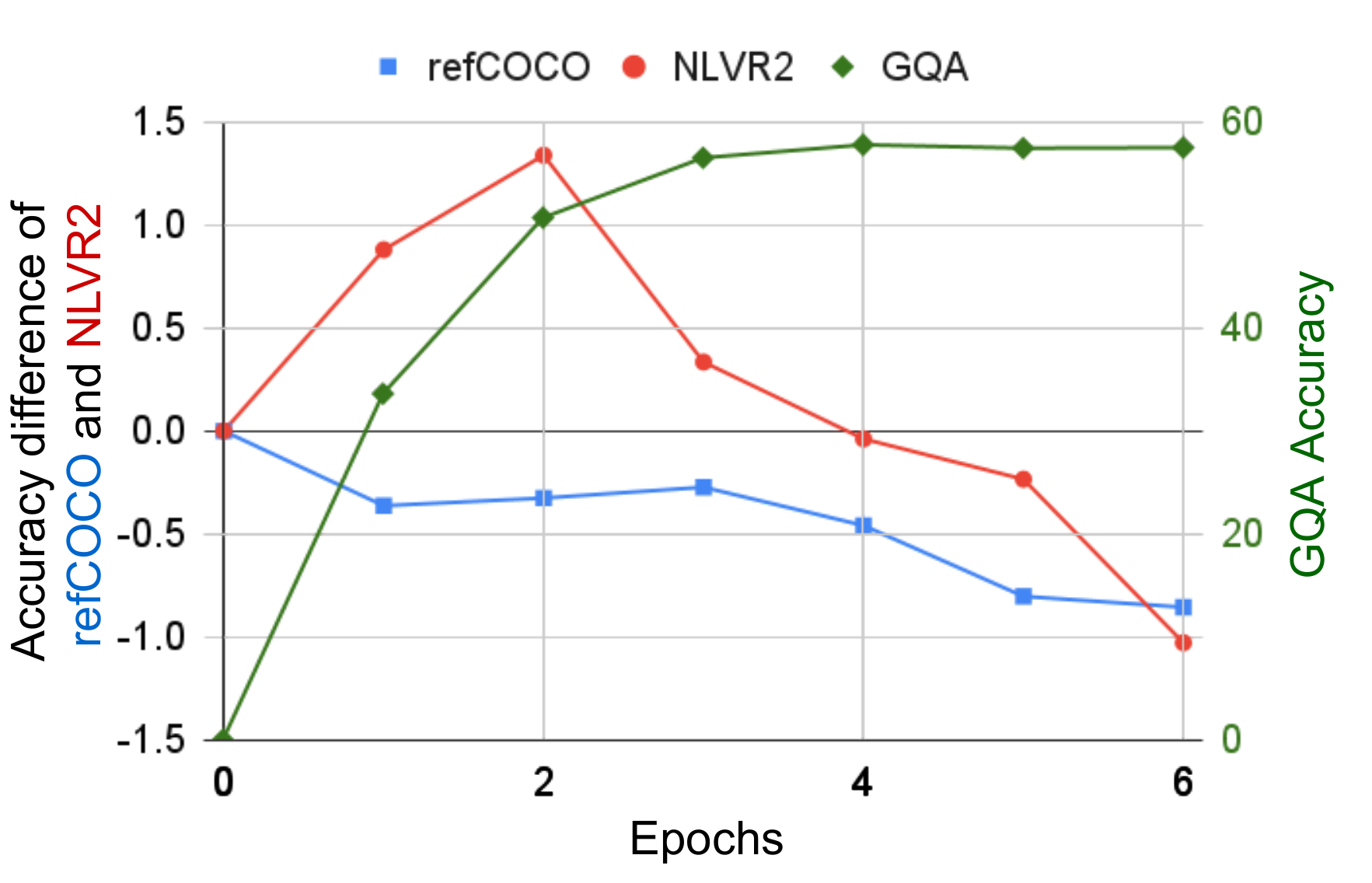}
    \caption{Accuracy on refCOCO ($\blacksquare$) and NLVR2 ($\bullet$) fine-tuned with $m_{\text{GQA}}$ after different epochs of pre-training. As a reference, the accuracy of GQA ($\blacklozenge$) is also shown.}
    \label{fig:result_ext_GQA_epochs}
\end{figure}

\begin{figure*}[t]
    \centering
    \includegraphics[width=0.95\linewidth]{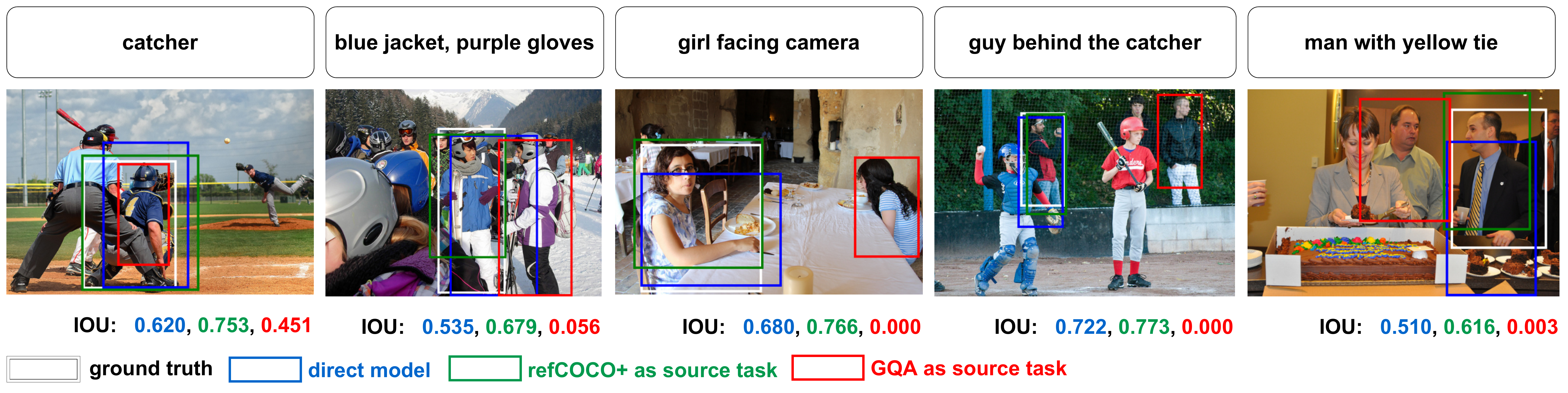}
    \caption{Example of the results on refCOCO. With the caption on top of the image, different model find different region on the image. It is easy to see that refCOCO+ help refCOCO to get more accurate prediction, while GQA misleading refCOCO to some wrong regions.}
    \label{fig:vis_refCOCO}
\end{figure*}
\begin{figure*}[ht]
\vspace{5pt}
    \centering
    \includegraphics[width=0.95\linewidth]{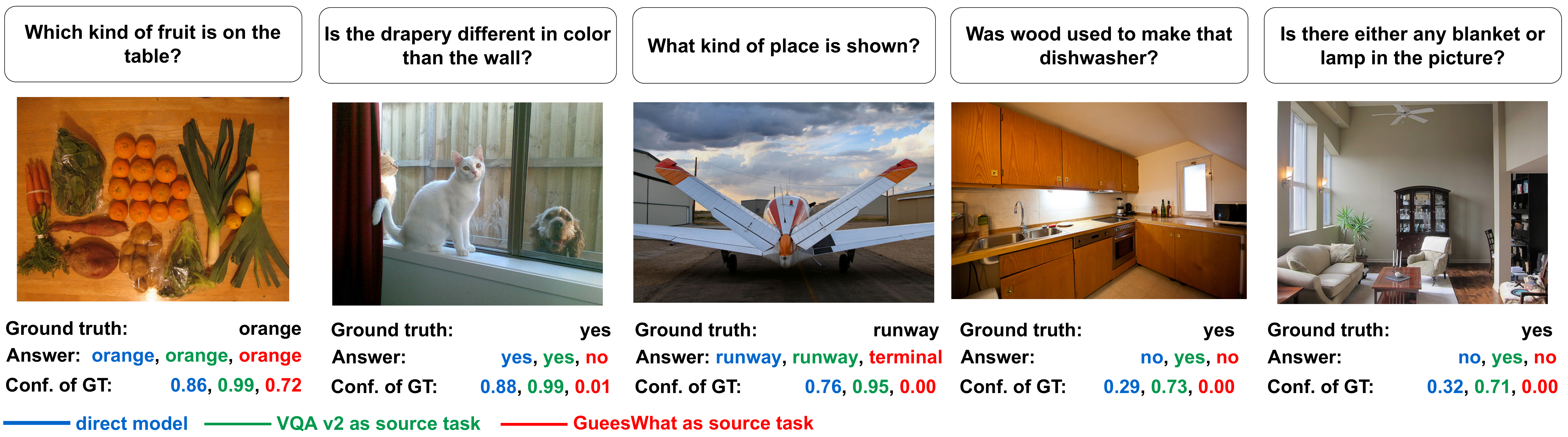}
    \caption{Example of the results on GQA. With the question on top of the image, different models predict the answer based on the image. The predictions from $m_{\text{GQA}}$, $m_{\text{VQA v2} \to \text{GQA}}$, and $m_{\text{GuessWhat} \to \text{GQA}}$, as well as the confidence score of the ground truth class (Conf. of GT), are shown under the examples respectively. It is easy to see that VQA v2 helps GQA to get a more accurate prediction, while GuessWhat misleading GQA to get a low confidence score in the ground truth class.}
    \label{fig:vis_GQA}
\end{figure*}

We finally explore the relationship between the number of training epochs of the source task and the success of the knowledge transferred in the target task. We conjecture that, for a target task that receives a negative effect from a source task, the more a model learns from the source task, the worse the model performs on the target task.
To verify this, we use GQA as the source task, which tends to give the most negative effect to the other tasks. 
Furthermore, we choose refCOCO and NLVR2 as the target tasks, which get the worst performance from GQA. 
We get six pre-trained $m_{\text{GQA}}^{e}$ models, where the number of epoch is ${e = \{1, \cdots,6\}}$. The higher the epoch, the more knowledge from GQA $m_{\text{GQA}}^{e}$ learns. We transfer these models to refCOCO and NLVR2. 

The results are shown in Figure \ref{fig:result_ext_GQA_epochs}. The blue $\blacksquare$ and red $\bullet$ are the scores for refCOCO and NLVR2, respectively, which for visibility are shown as the difference with respect to the direct model, \ie $a = {\text{Acc}_{\text{GQA} \to t}^{(e)} - \text{Acc}_{\text{GQA}}}$, where $\text{Acc}_{\text{GQA} \to t}^{(e)}$ is the accuracy of model pre-trained with GQA for $e$ epochs and fine-tuned with task $t$; $\text{Acc}_{\text{GQA} \to t}^{(0)}$ is the model that has no training on GQA, \ie the direct model; and $t$ is either refCOCO and NLVR2. 
For comparison, we also show the GQA accuracy ($\text{Acc}_\text{GQA}^{(e)}$, green $\blacklozenge$). 
Both tasks get lower score than the direct model when using a model trained on GQA for more than four epochs. In the case of refCOCO, it gets an inferior performance in all training epochs. In contrast,  
NLVR2 is improved by more than 1\% from GQA pre-trained for two epochs,  
showing that the knowledge from GQA not always has a negative effect.

\subsection{Visual results}

Figure \ref{fig:vis_refCOCO} shows predictions on refCOCO with the direct model $m_\text{refCOCO}$, one-hop model $m_{\text{refCOCO+} \to \text{refCOCO}}$, one-hop model $m_{\text{GQA} \to \text{refCOCO}}$, and the ground truth. 
The IOU between the prediction and the ground truth is shown under the respective image. 
Whereas refCOCO+ helps to find more accurate regions and to obtain higher IOU, GQA misleads the task to smaller or even wrong regions. For example, in the image in the middle, although the direct model finds the region with the right person, refCOCO+ helps to find a more accurate region but GQA predicts the wrong person. The same behavior can be observed in the last two images.

Figure \ref{fig:vis_GQA} shows examples on the GQA validation set with the direct model $m_{GQA}$, one-hop model $m_{\text{VQA\ v2} \to \text{GQA}}$, and one-hop model $m_{\text{GuessWhat} \to \text{GQA}}$.  We show the confidence of prediction for the ground truth class (Conf.~of GT) under each example. 
VQA v2 gives the most positive effect to GQA, while GuessWhat gives the most negative effect.
For example, in the second and third images from the left, GuessWhat induces to wrong answers, whereas in the last two images, VQA v2 helps to find the correct answers and improve the prediction with respect to the direct model.
More examples can be found in the Appendix.

\section{Conclusion}
We studied the knowledge transferability among 12 vision-and-language tasks. 
We confirmed that different tasks have different effect on each other, and the selection of tasks for knowledge transfer should be made carefully. 
In general, we observed some interesting insights about knowledge transferability, \eg, the tasks in the image retrieval and referring expressions groups tend to have a positive impact on other tasks, while the tasks in the visual question answering and multi-modal verification group give a negative contribution. The scale of datasets and the difference in their goals may cause this divergence.

\section{Acknowledgment}
This work was partly supported by JST CREST Grant No.~JPMJCR20D3, by JST FOREST Grant No.~JPMJFR216O, and by JSPS KAKENHI No.~JP22K12091.

{\small
\bibliographystyle{ieee_fullname}
\bibliography{main}
}

\begin{table*}[t]
    \centering
    \caption{Dataset statistics for the 12 tasks used in our experiments. From the left, the first and second columns are the number of samples in the train and validation (Train +Val) set and test set, respectively. The third column is the metric to evaluate the corresponding task. The fourth column is the name of the test set. The fifth column is the number of images in the train and validation set. The last column is the source dataset that the images of the corresponding task come from.
    }
    \renewcommand{\arraystretch}{1.2}
    \resizebox{!}{73pt}{
        \begin{tabular}{lrrllrl}
            \hline
            \multicolumn{1}{l}{}  & Train + Val samples & Test samples & Evaluation metric & Evaluation set & Train + Val image & Image source      \\ \hline
            VQA v2~\cite{VQA2.0}            & 542,104             & 447,793      & Accuracy          & test-dev                 & 98,861       & MSCOCO~\cite{COCOIR}            \\ 
            VG QA~\cite{VGQA}            & 1,294,255           & 5,000        & Accuracy          & validation                  & 92,147     & MSCOCO~\cite{COCOIR} + YFCC100M~\cite{YFCC100MTN} \\
            GQA~\cite{GQA}            & 962,928             & 12,578       & Accuracy          & test-dev                    & 69,868       & Visual Genome~\cite{VGQA}     \\ 
            COCO IR~\cite{COCOIR}            & 487,600             & 1,000        & Recall@5          & test                & 99,435           & MSCOCO~\cite{COCOIR}            \\ 
            Flickr30K IR~\cite{FlickrIR}            & 140,485             & 1,000        & Recall@5          & test           & 29,077           & Flickr30K~\cite{Flickr30K}         \\ 
            NLVR2~\cite{NLVR2}            & 86,373              & 6,967        & Accuracy          & test-P                  & 29,808         & NLVR2~\cite{NLVR2}             \\ 
            SNLI-VE~\cite{SNLIVE}            & 512,396             & 17,901       & Accuracy          & test                & 95,522           & Flickr30K~\cite{Flickr30K}         \\ 
            Visual7w~\cite{Visual7w}            & 93,813              & 57,265       & Accuracy          & test               & 16,415           & MSCOCO~\cite{COCOIR}            \\ 
            GuessWhat~\cite{GuessWhat}            & 100,398             & 23,785       & Accuracy          & test              & 51,291           & MSCOCO~\cite{COCOIR}            \\ 
            refCOCO~\cite{refCOCO1}            & 96,221              & 10,752       & Accuracy          & test                & 14,481           & MSCOCO~\cite{COCOIR}            \\ 
            refCOCO+~\cite{refCOCO1}            & 95,852              & 10,615       & Accuracy          & test               & 14,479           & MSCOCO~\cite{COCOIR}            \\ 
            refCOCOg~\cite{refCOCO2}            & 65,514              & 9,602        & Accuracy          & test               & 17,903           & MSCOCO~\cite{COCOIR}            \\ \hline
        \end{tabular}
        }
    \label{table:dataset_feature}
\end{table*}

\newpage

\appendix

{\Large{\textbf{Supplementary Material}}}

\section{Datasets statistics}

When training on multiple tasks, the overlapping between datasets may cause a data leak problem, e.g. images in refCOCO's test set also appear in COCO IR's train set. To avoid this, all of the training data is cleaned in \cite{12in1}. Please note that no cleaning process is done to the test set.
The statistics of the 12 dataset used in our experiments are shown in Table~\ref{table:dataset_feature}.

\section{More visual results}
\begin{figure*}
    \centering
    \includegraphics[width=0.97\linewidth]{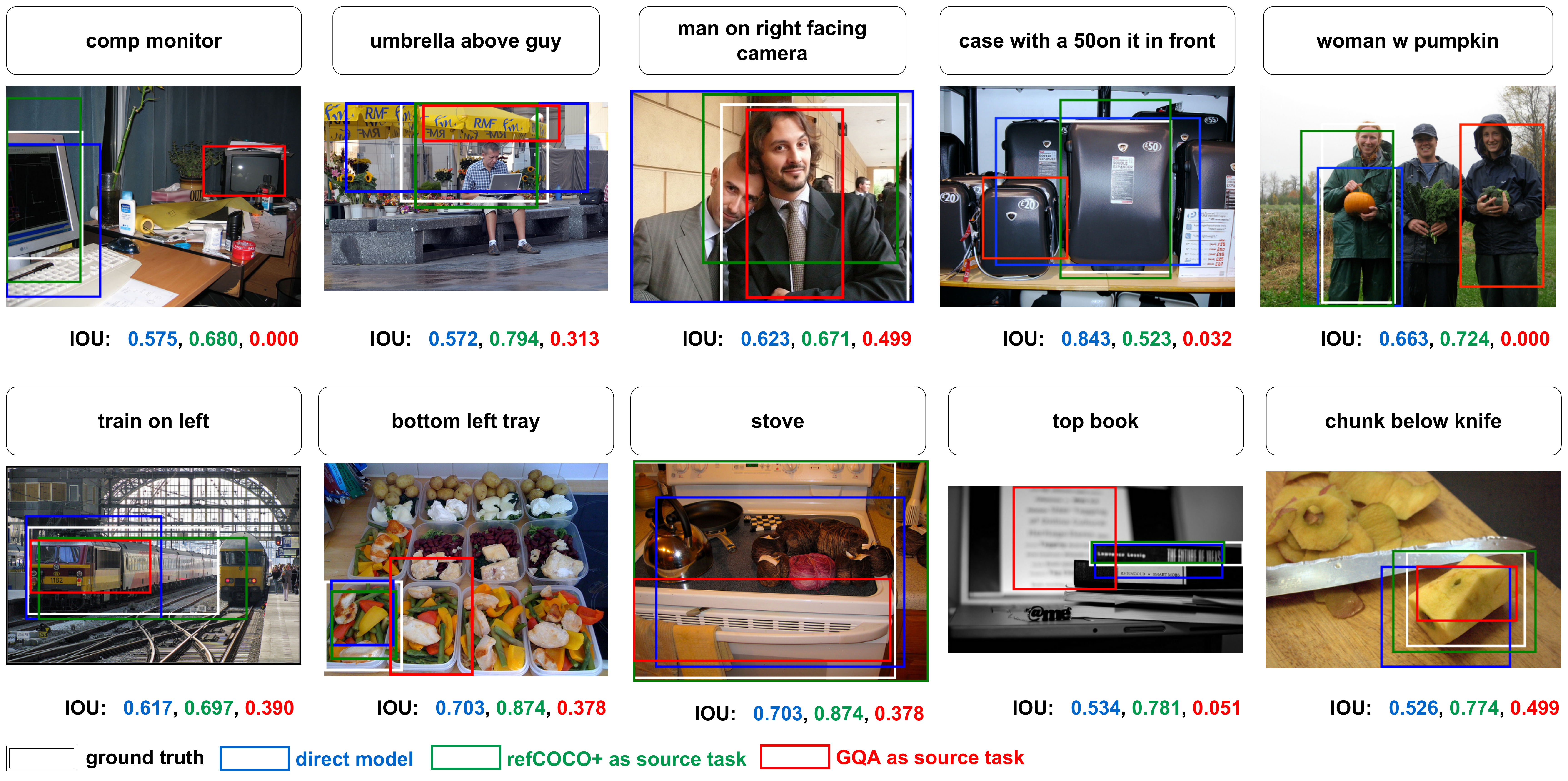}
    \caption{More examples of the results on refCOCO. With the caption on top of the image, different model find different region on the image. It is easy to see that refCOCO+ help refCOCO to get more accurate prediction, while GQA misleading refCOCO to some wrong regions.}
    \label{fig:vis_refCOCO_sup}
\end{figure*}

\begin{figure*}
\vspace{5pt}
    \centering
    \includegraphics[width=0.97\linewidth]{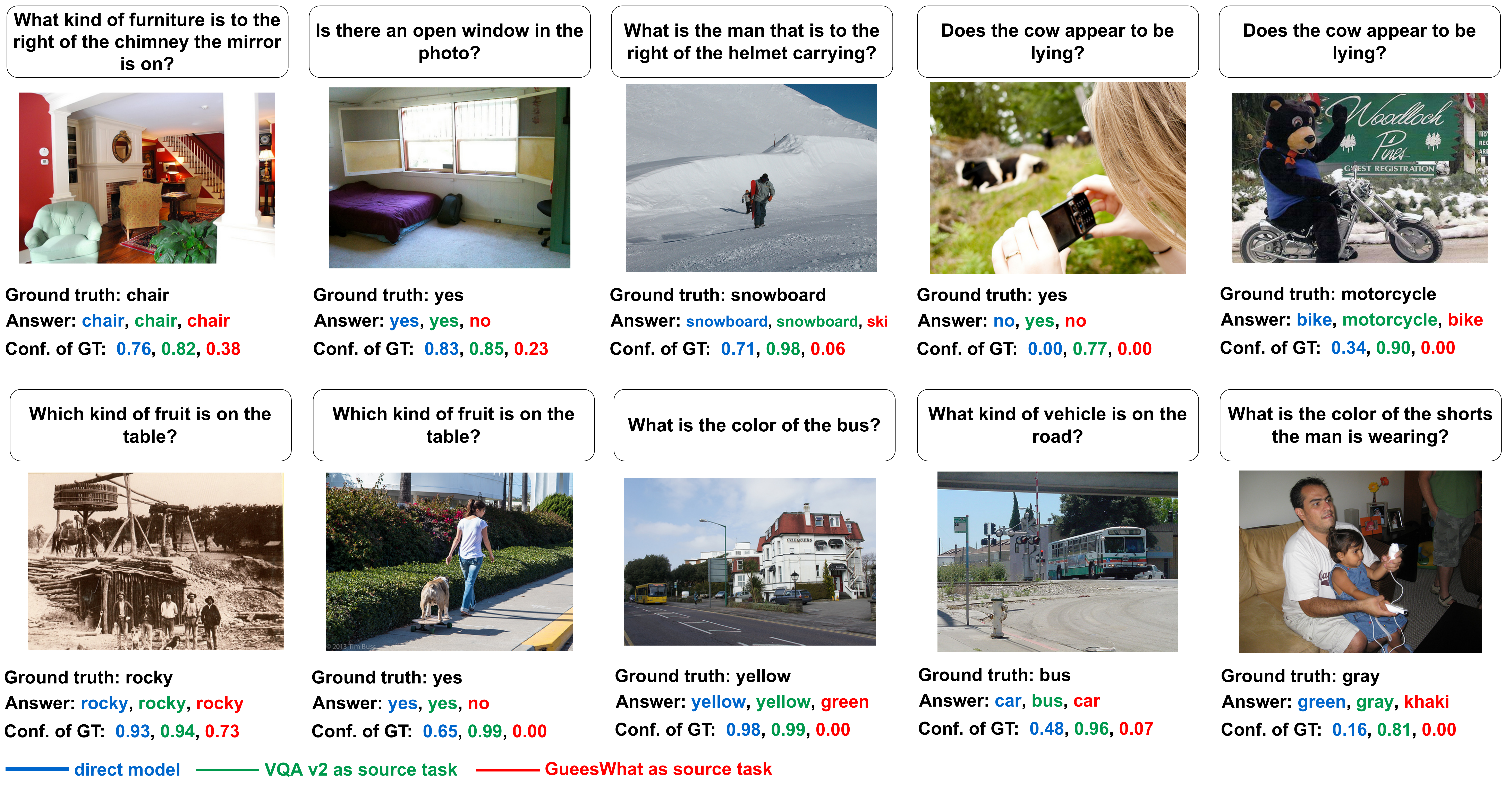}
    \caption{More examples of the results on GQA. With the question on top of the image, different models predict the answer based on the image. The predictions from $m_{\text{GQA}}$, $m_{\text{VQA v2} \to \text{GQA}}$, and $m_{\text{GuessWhat} \to \text{GQA}}$, as well as the confidence score of the ground truth class (Conf. of GT), are shown under the examples respectively. It is easy to see that VQA v2 helps GQA to get a more accurate prediction, while GuessWhat misleading GQA to get a low confidence score in the ground truth class.}
    \label{fig:vis_GQA_sup}
\end{figure*}

Figure~\ref{fig:vis_refCOCO_sup} shows additional 10 examples in refCOCO, and Figure~\ref{fig:vis_GQA_sup} shows 10 additional examples in GQA.

\end{document}